\pdfoutput=1
\newcommand{\CLASSINPUTtoptextmargin}{0.75in}
\newcommand{\CLASSINPUTbottomtextmargin}{0.75in}
\newcommand{\CLASSINPUTinnersidemargin}{0.75in}
\newcommand{\CLASSINPUToutersidemargin}{0.75in}
\documentclass[letterpaper,conference,10pt]{IEEEtran}
\IEEEoverridecommandlockouts
\usepackage[T1]{fontenc}

\usepackage{multicol}
\usepackage{hyperref}
\hypersetup{bookmarksopen,bookmarksnumbered,
pdfpagemode=UseOutlines,
colorlinks=true,
linkcolor=blue,
anchorcolor=blue,
citecolor=blue,
filecolor=blue,
menucolor=blue,
urlcolor=blue
}
\usepackage{gensymb}
\usepackage{amsmath, amssymb, amsfonts, amsthm}
\usepackage{bm}
\usepackage{subcaption}
\usepackage{flushend}
\usepackage{graphicx}
\usepackage{mathtools}
\usepackage[sort&compress,numbers]{natbib}
\let\degree\relax

\usepackage[utf8]{inputenc} 
\usepackage[T1]{fontenc}    
\usepackage{hyperref}       
\usepackage{url}            
\usepackage{booktabs}       
\usepackage{amsfonts}       
\usepackage{nicefrac}       
\usepackage{microtype}      

\usepackage[utf8]{inputenc}
\usepackage{indentfirst}
\usepackage{amsmath}
\usepackage{graphicx}
\usepackage{amssymb}
\usepackage{bm} 
\usepackage{mathtools}

\usepackage{algorithm}
\usepackage{caption}
\usepackage{subcaption}
\usepackage{graphicx}
\usepackage{algcompatible}
\usepackage{booktabs}

\DeclareMathOperator*{\argmin}{arg\,min}

\newcommand{\vi}[1]{\mathbf{#1}}

\newcommand{\prob}{\vi{p}}
\newcommand{\state}{\vi{x}}
\newcommand{\nominalstate}{\bar{\vi{x}}}
\newcommand{\nominalcontrol}{\bar{\vi{u}}}
\newcommand{\dstate}{\delta\vi{x}}
\newcommand{\action}{a}
\newcommand{\control}{\vi{u}}
\newcommand{\dcontrol}{\delta\vi{u}}
\newcommand{\cost}{{l}}

\newcommand{\Quu}{\vi{Q}_{\control\control}}
\newcommand{\Qu}{\vi{Q}_{\control}}
\newcommand{\Qux}{\vi{Q}_{\control\state}}
\usepackage[font=footnotesize]{caption} 
\usepackage[font=footnotesize]{subcaption}
\usepackage{mathabx}
\usepackage{verbatim}
\usepackage{algorithm}
\usepackage[noend]{algpseudocode}

\usepackage{multirow}
\usepackage{booktabs}
\usepackage[dvipsnames]{xcolor}

\usepackage[]{color}

\makeatletter
\def\BState{\State\hskip-\ALG@thistlm}
\makeatother

\begin{document}

\title{\vspace{0.20in}\LARGE Hybrid DDP in Clutter (CHDDP): Trajectory Optimization for Hybrid Dynamical System in Cluttered Environments}
\author
{
\IEEEauthorblockN{Shushman Choudhury} \IEEEauthorblockA{Stanford University \\ shushman@stanford.edu} \and
\IEEEauthorblockN{Yifan Hou} \IEEEauthorblockA{Carnegie Mellon University \\ yifanh@andrew.cmu.edu}
\and
\IEEEauthorblockN{Gilwoo Lee, Siddhartha S. Srinivasa} \IEEEauthorblockA{University of Washington \\ \{gilwool,siddh\}@cs.uw.edu}
}

\maketitle

\begin{abstract}
We present an algorithm for obtaining an optimal control policy for hybrid dynamical systems in cluttered environments. To the best of our knowledge, this is the first attempt to have a locally optimal solution for this specific problem setting. Our approach extends an optimal control algorithm for hybrid dynamical systems in the obstacle-free case to environments with obstacles. Our method does not require any preset mode sequence or heuristics to prune the exponential search of mode sequences. By first solving the relaxed problem of getting an obstacle-free, dynamically feasible trajectory and then solving for both \emph{obstacle-avoidance} and \emph{optimality}, we can generate smooth, locally optimal control policies. We demonstrate the performance of our algorithm on a box-pushing example in a number of environments 
against the baseline of randomly sampling modes and actions with a Kinodynamic RRT.
\end{abstract}

\IEEEpeerreviewmaketitle
\section{Introduction}
\label{sec:intro}

We consider the problem of controlling a hybrid system with discrete modes and continuous inputs in a cluttered environment. Many robotics systems for real-world applications are hybrid in nature. For example, while operating an autonomous car, the system must choose an appropriate (discrete) gear and control the (continuous) acceleration. When pushing a block on a surface, the end-effector must choose which (discrete) face to push, with each face having its own (continuous) dynamics. Obstacle-avoidance must also be ensured in both of these cases.

Optimal control of hybrid systems is a crucial aspect of robotic navigation and manipulation. The problem characteristics make it quite difficult to solve efficiently. The discreteness of modes makes the optimization problem non-convex, and makes the control input space exponential in the length of the planning horizon. The presence of obstacles, on the other hand, makes the feasible state-space non-convex.

Although many robotic systems have hybrid nature, our specific problem has not been considered previously in the literature. Recently, the optimal hybrid control problem in the obstacle-free setting, was solved using Differential Dynamic Programming (DDP)~\citep{pajarinen2017hybrid}. They relaxed the hybrid constraint and transformed the discrete controls into a convex combination of weights. We discuss other related work in Section~\ref{sec:related}, and additional background material in Section~\ref{sec:prelims}. 

Our key insight is that locally optimal methods for hybrid systems in clutter often get stuck in poor local minima because simultaneously finding a feasible mode sequence and moving away from the obstacles is challenging. We address these two challenges by decoupling them and solving the problem in two stages: 1) we generate a \emph{collision-free} and  \emph{dynamically feasible} trajectory and then 2) we solve for a \emph{collision-free}, \emph{optimal} policy using the mode sequence found in the first stage. We describe our approach in detail in Section~\ref{sec:approach}. 

\begin{figure}[t!]
    \begin{subfigure}[b]{0.49\columnwidth}
    \centering
    \includegraphics[trim={2.5cm 1.2cm 2.5cm 0.8cm},clip,width=\textwidth]{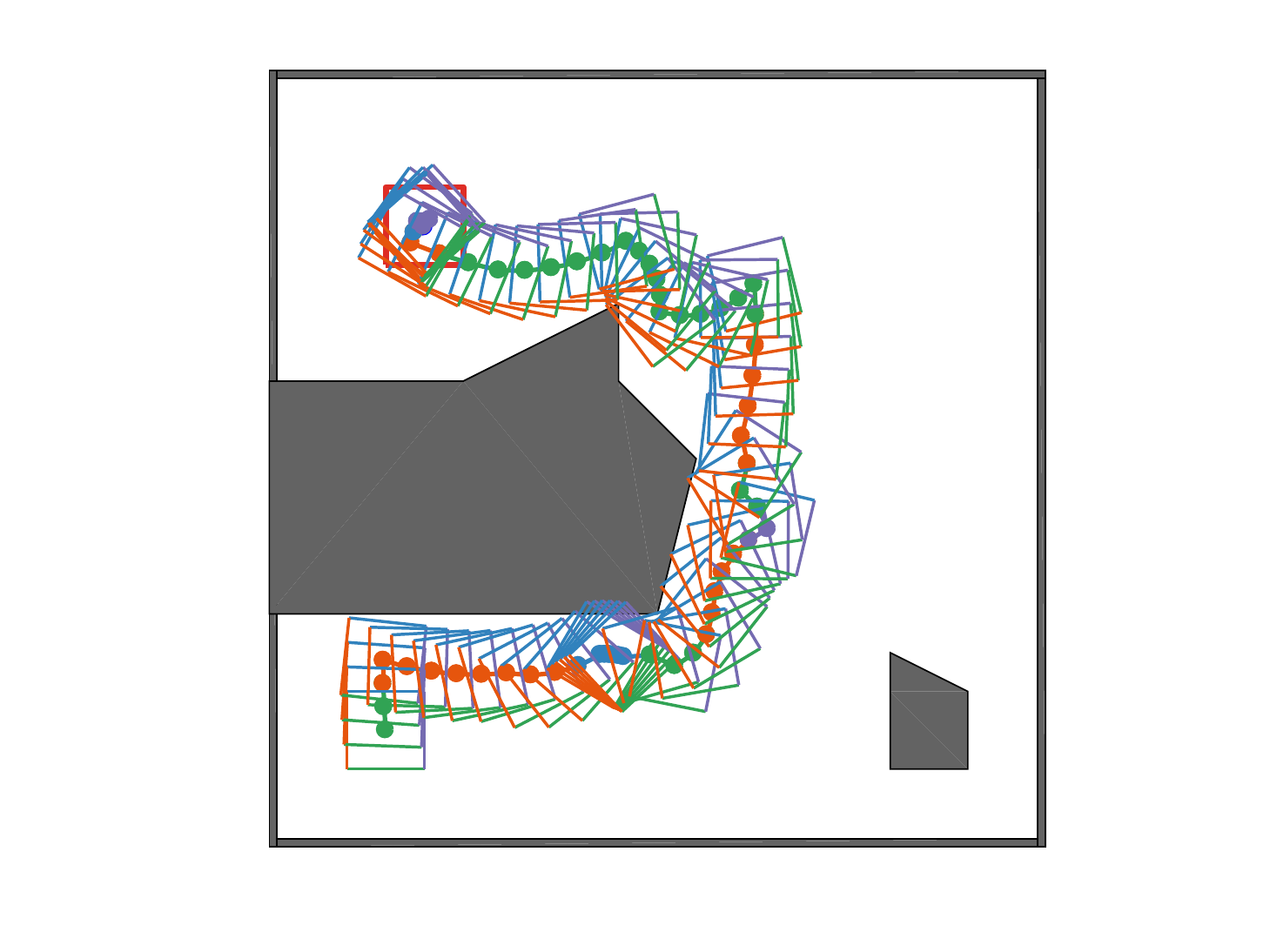}
    \caption{KDRRT}
    \label{fig:w2_rrt}
    \end{subfigure}
    \begin{subfigure}[b]{0.49\columnwidth}
    \centering
    \includegraphics[trim={2.5cm 1.2cm 2.5cm 0.8cm},clip,width=\textwidth]{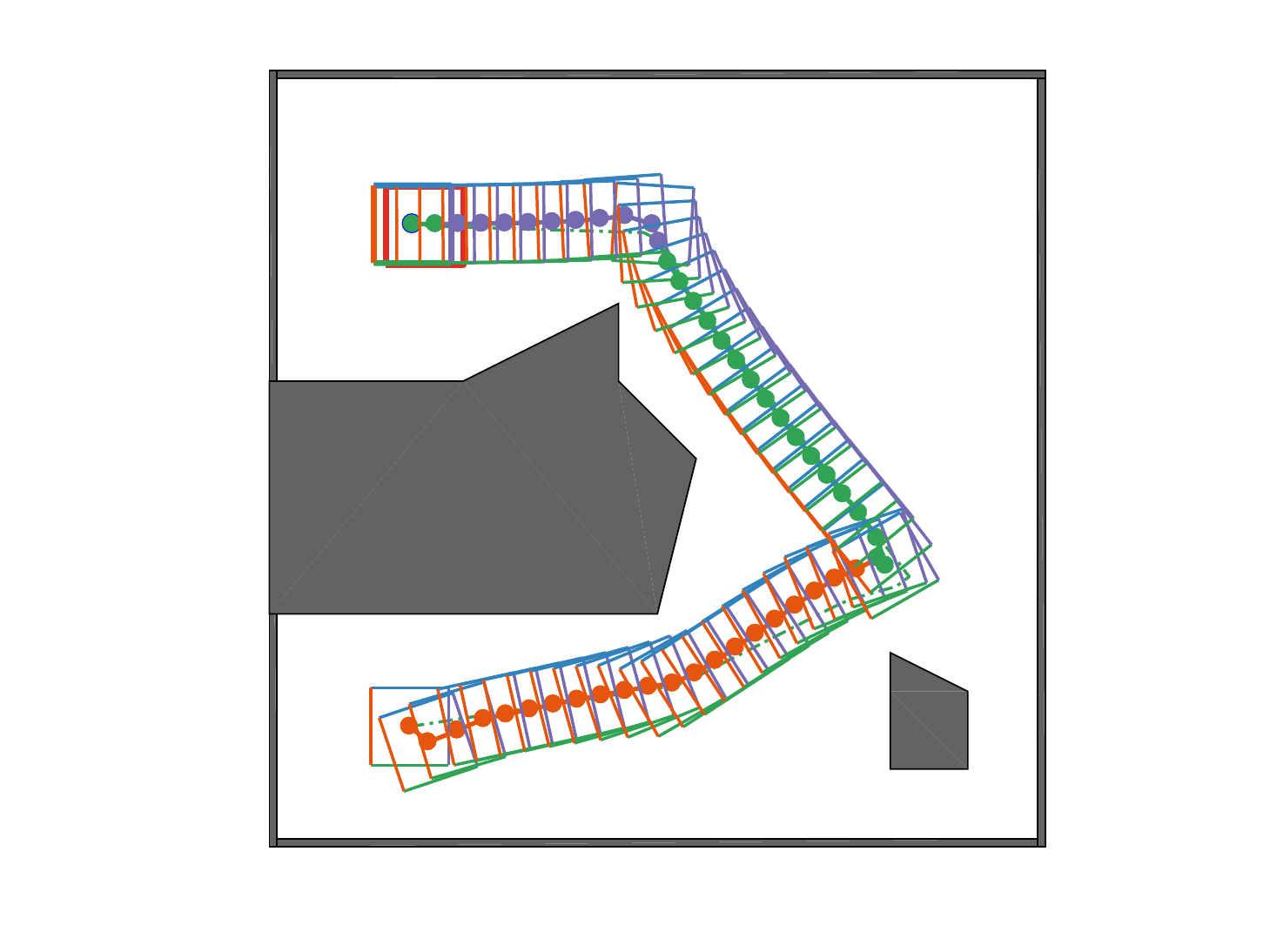}
    \caption{CHDDP}
    \label{fig:w2_hddp}
    \end{subfigure}
    \caption{Our approach (CHDDP) obtains a locally optimal state-action-mode trajectory and control policy for a hybrid system, in this case a box being pushed. The colour of each trajectory segment represents a particular mode, each of which corresponds to a side of the box that the pusher acts on. The trajectory obtained by CHDDP is significantly smoother than that obtained by the Kinodynamic RRT and has fewer mode switches.}
    \label{fig:w2_homotopy}
\end{figure}

Our major contribution is the ability to produce locally optimal control policies for hybrid systems with both state and input constraints. To the best of our knowledge, our algorithm is the first approach for this specific problem domain. Our method does not require a mode sequence to be specified beforehand, nor does it require any heuristics to prune an exponential search space of discrete mode transitions. We achieve wider applicability for the previously introduced approach for hybrid control~\citep{pajarinen2017hybrid} by non-trivially extending it to handle the presence of clutter. 

We compare our approach CHDDP (Constrained Hybrid DDP) to the Kinodynamic RRT (KDRRT)~\citep{lavalle2001randomized} which
randomly samples modes and actions without any notion of solution quality or optimality. We show qualitative and quantitative results over several box-pushing
problems in Section~\ref{sec:results}. Finally, we conclude in Section~\ref{sec:conclusion}
with a discussion of some limitations and questions for future research.


\begin{figure*}[t]
    \centering
    \begin{subfigure}[b]{0.245\textwidth}
        \centering
        \includegraphics[width=\textwidth]{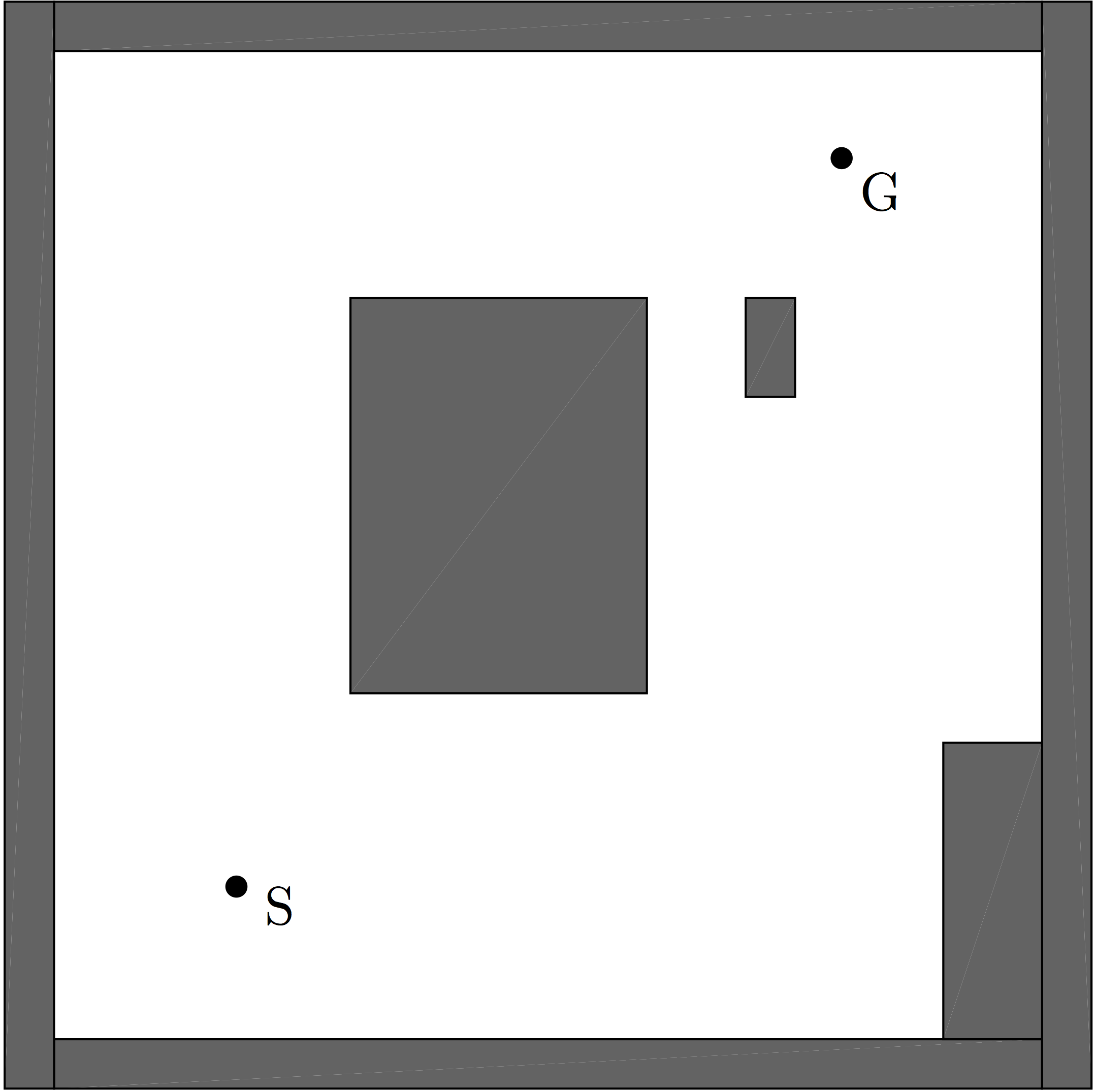}
        \caption{}
        \label{fig:expl_w7}
    \end{subfigure}
    \begin{subfigure}[b]{0.245\textwidth}
        \centering
        \includegraphics[width=\textwidth]{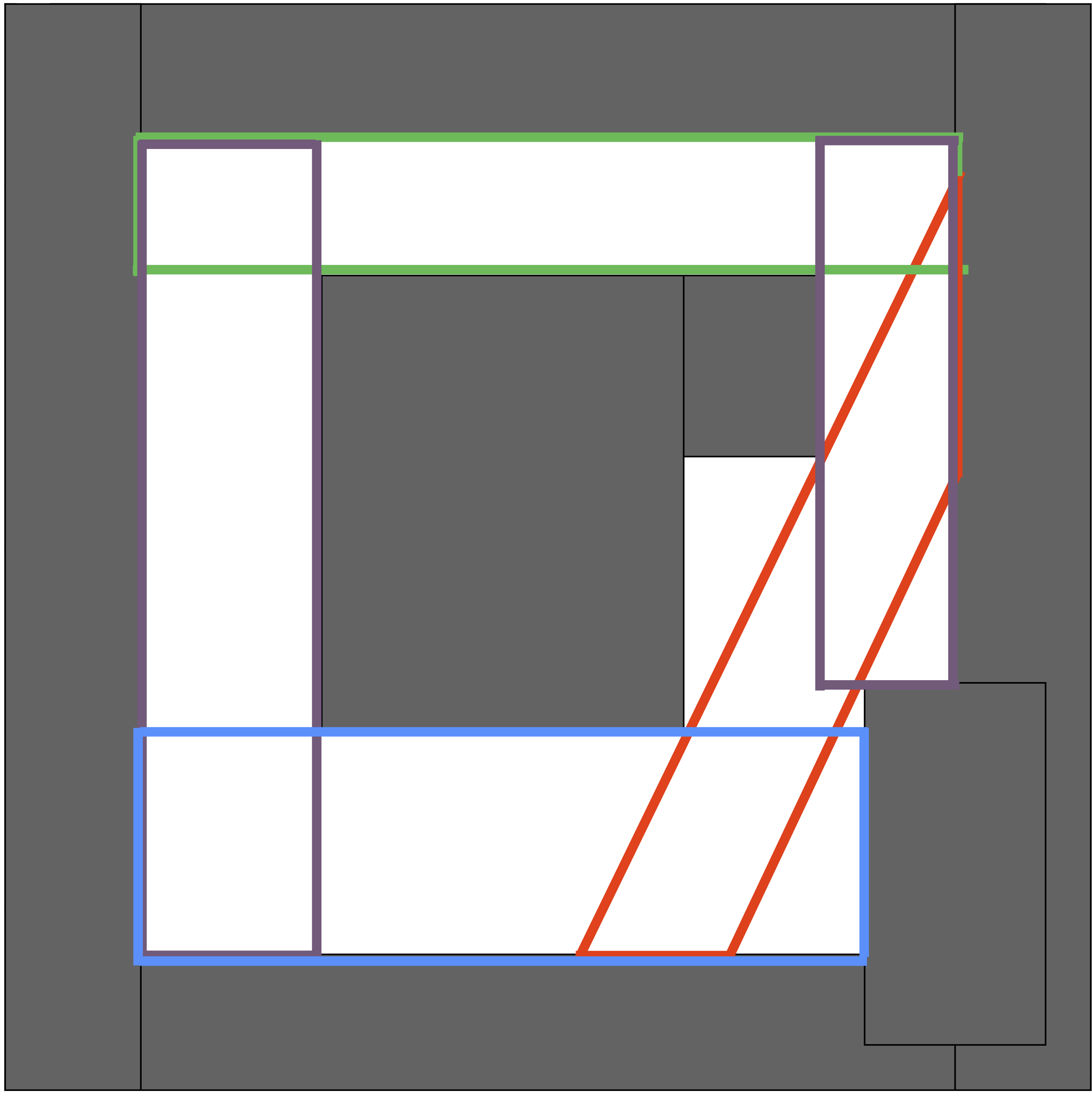}
        \caption{}
        \label{fig:expl_hyper}
    \end{subfigure}
    \begin{subfigure}[b]{0.245\textwidth}
        \centering
        \includegraphics[width=\textwidth]{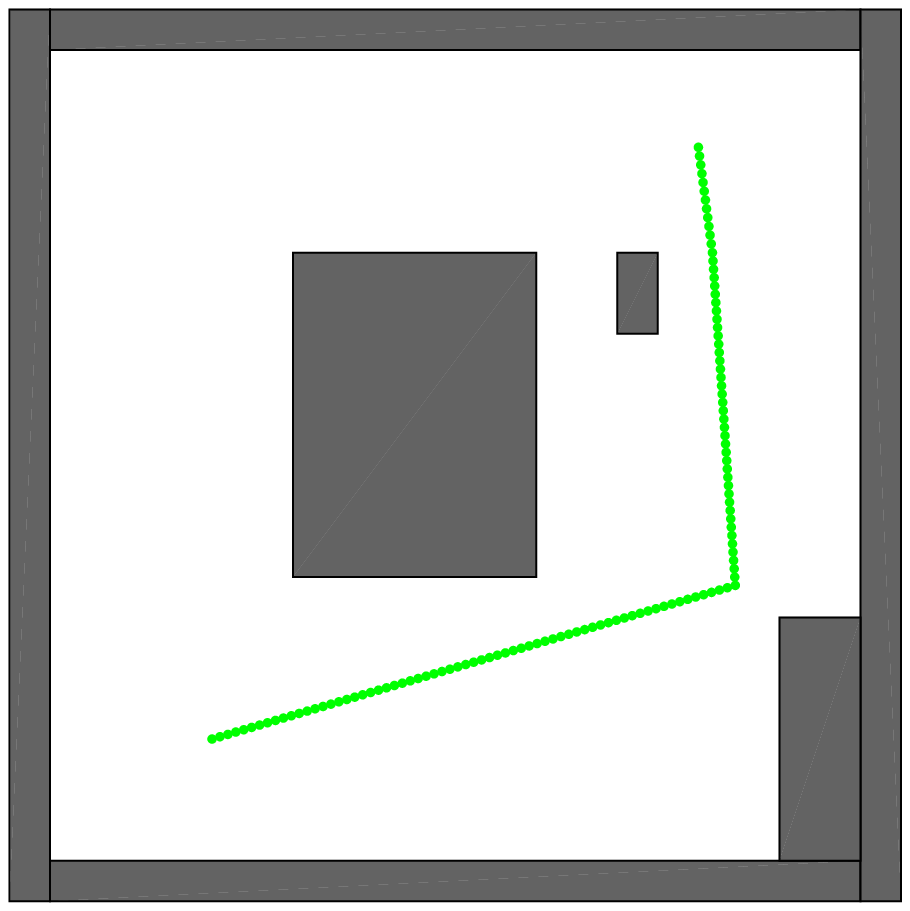}
        \caption{}
        \label{fig:expl_hom}
    \end{subfigure}
    \begin{subfigure}[b]{0.245\textwidth}
        \centering
        \includegraphics[width=\textwidth]{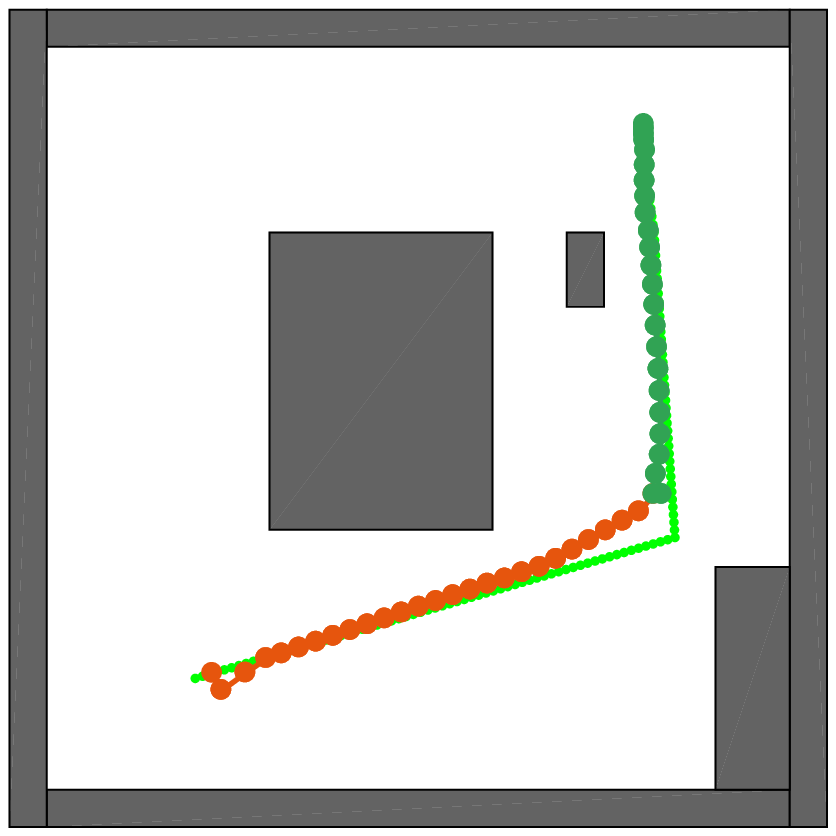}
        \caption{}
        \label{fig:expl_ddp}
    \end{subfigure}
    \caption{An overview of our approach. (\subref{fig:expl_w7}) The environment with desired start and goal coordinates,
    and obstacles. (\subref{fig:expl_hyper}) The feasible state space is computed as a union of convex polytopes, which
    simplifies the problem and makes collision-checking efficient. Note that we do this with inflated obstacles so that geometrically feasible paths have sufficient clearance for CHDDP to work well (\subref{fig:expl_hom}) A feasible nominal path, without mode
    sequences, which is used to initialize the next stage of CHDDP. (\subref{fig:expl_ddp}) A state-action-mode
    trajectory which closely tracks the initial solution and is the result of our overall approach. The colours along the trajectory represent the mode at each timestep.}
    \label{fig:expl}
\end{figure*}

\section{Related Work}
\label{sec:related}

The problem of obtaining locally optimal trajectories for a hybrid control system has been addressed in numerious previous works~\citep{branicky1998unified,nandola2008multiple,zhu2015optimal}, but incorporating obstacles is a highly non-trivial challenge. The feasible set of solutions is non-convex because of the obstacles. The objective function is non-convex because of the hybrid nature of the dynamics.

In the hybrid control community, Multiple Lyapunov Functions (MLF) have been used for control law synthesis on switched nonlinear systems \citep{el2005output}.  However, Lyapunov analysis framework is not designed for handling obstacle avoidance problems. 
Another thread of work on hybrid control synthesis is Mixed Integer Programming (MIP), which has been widely applied to locomotion \citep{deits14footstep}, UAVs \citep{richards2002coordination,deits2015efficient} and autonomous vehicles \citep{schouwenaars2001mixed}. However, the discreteness of modes makes any naive search of mode sequences grow exponentially. In order to solve MIP more efficiently, \emph{reduction} of search space or \emph{relaxation} of hybrid constraints has proved effective \citep{lincoln02, hogan2016feedback}.  Real-time feedback control has been produced by reducing the search space into a much smaller set of mode sequences~\citep{hogan2016feedback}. When dynamic constraints is present, shooting methods like DDP could converge faster than direct methods like MIP. Relaxing the hybrid constraint by representing discrete modes as a probability distribution has enabled shooting methods to solve hybrid problem without requiring heuristics~\citep{pajarinen2017hybrid}.

In addition to the hybrid nature of the dynamics system, the existence of obstacles makes the feasible state-space non-convex. Trajectory planning methods based on optimization~\citep{ratliff2009chomp, schulman2014motion} are good at handling obstacles, although the lack of a steering function for hybrid dynamic systems makes it challenging to extend these approaches to the hybrid domain. Sampling-based planning methods such as the kinodynamic RRT~\citep{lavalle2001randomized} can handle both hybridness and clutter, but the quality of solutions can be arbitrarily poor and they are only open-loop policies as opposed to closed-loop feedback laws. 

\section{Preliminaries}
\label{sec:prelims}

\begin{figure*}[th]
    \centering
    \begin{subfigure}[b]{0.325\textwidth}
        \centering
        \includegraphics[trim={2.5cm 1.2cm 2.5cm 0.8cm},clip,width=\textwidth]{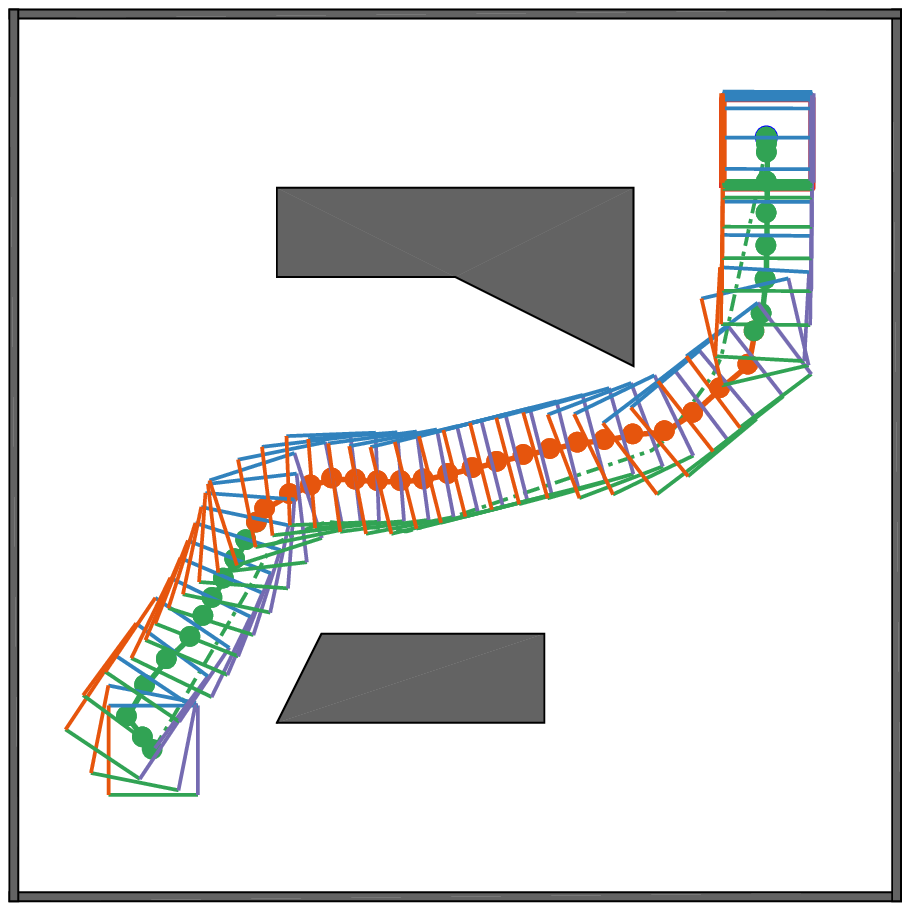}
        \caption{}
        \label{fig:w5_hom1}
    \end{subfigure}
    \begin{subfigure}[b]{0.325\textwidth}
        \centering
        \includegraphics[trim={2.5cm 1.2cm 2.5cm 0.8cm},clip,width=\textwidth]{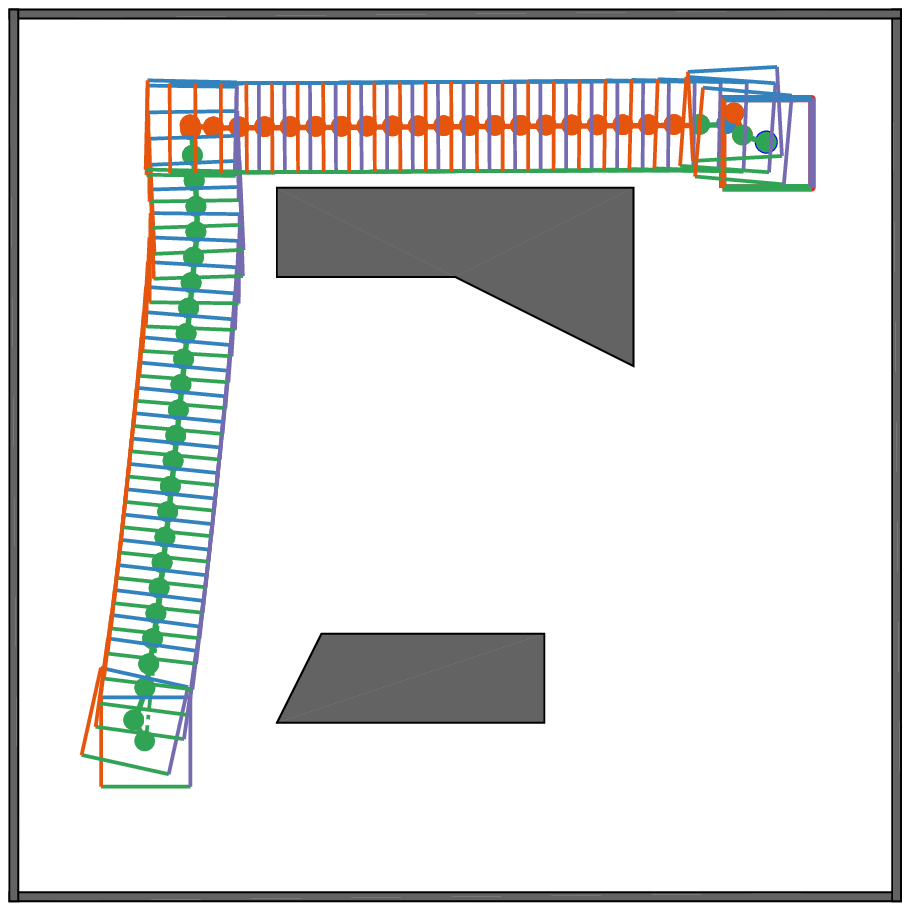}
        \caption{}
        \label{fig:w5_hom2}
    \end{subfigure}
    \begin{subfigure}[b]{0.325\textwidth}
        \centering
        \includegraphics[trim={2.5cm 1.2cm 2.5cm 0.8cm},clip,width=\textwidth]{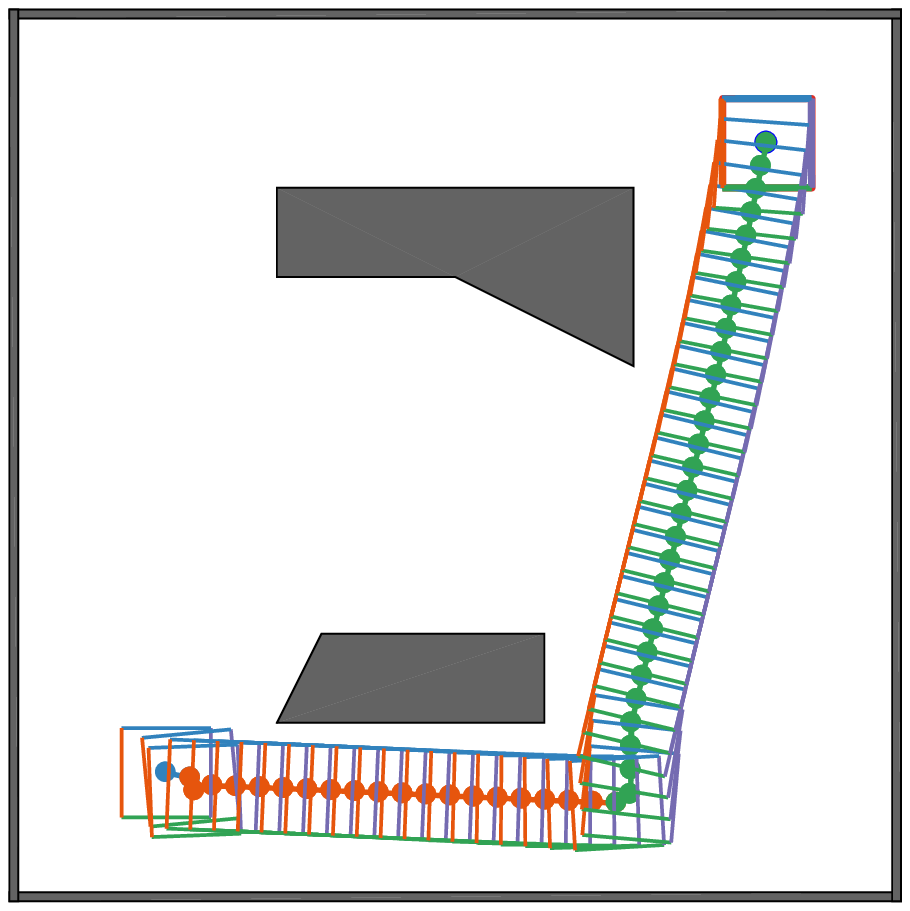}
        \caption{}
        \label{fig:w5_hom3}
    \end{subfigure}
    \caption{For generating the initial feasible path required by CHDDP in problems that can be projected to 2D, we use a heuristic method based on visibility graphs
    (with inflated obstacles) to get paths in multiple homotopy classes. This method is reasonable as it returns short paths that have good clearance and cover multiple homotopy classes, if applicable. It becomes particularly useful in case one homotopy class is better suited for a locally optimal method like CHDDP. Shown here are solutions of CHDDP using initial paths from multiple homotopy classes. }
    \label{fig:w5_hom}
\end{figure*}

\subsection{Problem Statement}
\label{sec:problem}

 A hybrid dynamical system \citep{pajarinen2017hybrid} is a set of continuous dynamical systems indexed by discrete modes. We consider a subclass of such systems in which the mode transition can be made arbitrarily. Within each mode, the dynamics is defined as the following: 
\begin{equation}\label{eq:dynamic}
    \state_{t+1} = f_{\action_t}(\state_t, \control_t)
\end{equation}
where $t$ is the current timestep, $\action_t$ is the current mode, ${\control}_t \in U$ is the current control input, ${\state_t}\in X$ is the continuous state of the system, and $f_{\action_t}: X \times U \rightarrow X$ is a continuous function that represents the dynamics of the current mode. The set of all dynamics functions is denoted as $F$. Each mode $\action_t$ is chosen among $N_a$ possible modes and has bounded control inputs, i.e. ${\control}_t \in [\underline{\action_t}, \overline{\action_t}] $.
The configuration space $\mathcal{X} \subset X$ is a subspace of $X$ which represents the system kinematics, e.g. rigid body pose, or joint angles for a manipulator. The set of all configurations in collision with the environment and in self-collision (if applicable) is denoted as $\mathcal{X}_{obs}$ and the complement of this is the free space $\mathcal{X}_{free} = \mathcal{X} \setminus \mathcal{X}_{obs}$. When we say $\state_t \in \mathcal{X}_{free}$ or
$\state_t \in \mathcal{X}_{obs}$, we mean that the configuration embedded in $\state_t$ 
is in $\mathcal{X}_{free}$ or $\mathcal{X}_{obs}$.

A trajectory $\tau$ is defined as a sequence of state-mode-control tuples $\{(\state_t,\action_t,\control_t)\}_{t=1}^T$. Given a hybrid dynamical system and an environment with obstacles, our objective is to find a locally optimal trajectory $\tau^*$ that minimizes the total cost
\begin{align}\label{eq:objective}
\tau^* = \argmin_{\tau} ~ \cost_T(\state_T) + \sum_{t=0}^{T-1}l_{\action_t}(\state_t, \control_t)
\end{align}
subject to dynamics, collision constraints and specified start and goal states
$\state_s$ and $\state_f$ respectively.
Here, $\cost_{a_t}({\state}_t, {\control}_t)$ is the immediate cost for executing control ${u}_t$ under mode $a_t$, and  $\cost_T(\state_T)$ is the final cost.

\subsection{Differential Dynamic Programming}
 We briefly review DDP~\citep{jacobson1970differential,ControlLimitedDDP} and explain how we extend it to hybrid systems.

The optimal cost-to-go, a.k.a. \emph{value} of a state $\state$ at timestep $t$ can be found by recursively choosing actions that minimize the total cost:
%
\begin{equation}
    \label{eq:ddp bellman}
    V_t(x) = \min_\control \ \cost(\state,\control) + V_{t+1}(f(\state, \control)).
\end{equation}

Given a nominal trajectory $(\bar x, \bar u)$, let $Q$ be the increase of $V_t$ (\ref{eq:ddp bellman}) caused by a change of trajectory $(\delta x, \delta u)$:
\begin{align*}
    &Q(\dstate, \dcontrol) :=   l(\nominalstate + \dstate, \nominalcontrol + \dcontrol) + \\
    &V_{t+1}(f(\nominalstate + \dstate, \nominalcontrol + \dcontrol )) - \big(l(\nominalstate, \control) + V_{t+1}(f(\nominalstate, \nominalcontrol))\big)
\end{align*}
\noindent Then, a locally optimal change can be made by choosing $\delta u$ as the solution to the following problem:
\begin{equation}\label{eq:ddp-objective}
    \begin{array}{rl}
&\dcontrol^*= \mathop {\arg \min }\limits_{\dcontrol} Q(\dstate,\dcontrol)\\
 &\approx \mathop {\arg \min }\limits_{\delta u} \frac{1}{2}\dcontrol^\top{\Quu}\dcontrol+ \Qu^\top \dcontrol+ \dcontrol^\top \Qux\dcontrol.
\end{array}
\end{equation}
The second equation comes from a local quadratic approximation of $Q$, which gives the following solution:
\begin{align} \label{eq:local-linear}
    \dcontrol^* =  -\Quu^{-1} \Qu  -\Quu^{-1}\Qux\dstate
\end{align}
The first term is the changed nominal control and the second term is a linear feedback control law. $\Quu, \Qu, \Qux$ can be computed recursively from $V_T(\bar \state_T)$~\citep{jacobson1970differential} during a backward pass. DDP iterates between a backward pass and a forward rollout until the change in nominal control is small.

\subsection{DDP for Hybrid System (HDDP)}

\label{subsec:switched}
The classic DDP method can only solve smooth problems. It has been extended to solve switched hybrid control problems~\citep{pajarinen2017hybrid}, where the control $\hat u$ contains both continuous action $u$ and discrete action $a$. They replaced the discrete control with a continuous vector $\bf p$, each element of which represents the probability of choosing one specific discrete mode, i.e. $\bf{\hat u} = \begin{bmatrix} \bf{u} & \bf{p} \end{bmatrix}$, with $p_a \in [0, 1], \forall a$. Then they used DDP to solve the convexified, smooth dynamics: 
\begin{equation}
    \hat f(\state,{\bf{\hat \control}}) = \sum\limits_a {{p_a}f_a(\state,\control)}.
    \label{eq:dummy_dynamics}
\end{equation}
The following term is added to the cost function
\begin{equation*}
c_{ST}({\state,\control,\prob}) = C_{ST}\sum\limits_a 
\left\{
\begin{array}{*{20}{c}}
\phi ({p_a}) & \rm{if\ }{p_a} < p_{th}\\
\phi (\frac{1 - {p_a}}{p_{th}(1 - {p_{th}})}) &{\rm{if\ }{p_a} \ge p_{th}}
\end{array}
\right\}
\end{equation*}
where $\phi(\cdot)$ is the pseudo-Huber loss, $p_{th}=1/N_a$ and $C_{ST}$ is a coefficient that grows along with the number of DDP iterations. A large $C_{ST}$ drives each $p_i$ to either zero or one, so that when DDP converges, the solution to (\ref{eq:dummy_dynamics}) has a fixed sequence of modes, which is also a solution to the original hybrid problem (\ref{eq:dynamic}). 

Constraints on inputs, i.e. $\forall a, 0\le p_a \le 1$ and $\sum_a{p_a} = 1$ are handled \citep{ControlLimitedDDP} by solving Equation \ref{eq:ddp-objective} for $\dcontrol$  as a constrained quadratic program.


\section{Approach}
\label{sec:approach}

Extending obstacle-free hybrid control approaches~\citep{pajarinen2017hybrid,el2005output, johnson2016hybrid} to a cluttered setting is nontrivial. Our key insight is that locally optimal approaches for hybrid systems in clutter often get stuck in poor local minima because simultaneously finding a feasible mode sequence and moving away from obstacles is challenging. We address these two challenges by decoupling them and running DDP twice to solve the problem: 
\begin{enumerate}
\item We generate a \emph{collision-free geometric path} using some path-planning algorithm and obtain a \emph{dynamically feasible trajectory} using DDP without considering obstacles.
\item Using the mode sequence found from step 1, we solve for a \emph{collision-free}, \emph{locally optimal} policy using DDP with (efficient) collision checking.
\end{enumerate}
This decoupling is non-trivial and at the heart of our approach. We discuss it in detail here.

\subsection{ Generating a collision-free, dynamically feasible nominal trajectory}

During the DDP forward passes, whenever the state violates collision constraints, i.e. is in $\mathcal{X}_{obs}$, we project it back to $\mathcal{X}_{free}$, which requires many collision checks and could be computationally expensive. To reduce the computation, we approximate the collision-free space as a union of convex polytopes using IRIS (Iterative Regional Inflation by Semidefinite
programming)~\citep{deits2015computing} (See Figure \ref{fig:expl_hyper} for illustration). Instead of collision checking, the constraint check is performed by checking  whether a given point $\state$ belongs to one of the convex regions, and if not,  it returns the point projected to the closest convex region. Note that this union of convex polytopes is a conservative approximate of $\mathcal{X}_{free}$, which is often beneficial for our purpose as long as the space reduction is not too significant.

\begin{algorithm}[t]
\caption{Hybrid DDP in Clutter (CHDDP)} \label{alg:chddp}
\begin{algorithmic}[1]
\renewcommand{\algorithmicrequire} {\textbf{Input :} }
\renewcommand{\algorithmicensure} {\textbf{Output :} }
\Require $\mathbf{x}_s$, $\mathbf{x}_f$, $X$, $U$, $F$
\Statex $\xi \gets \texttt{Geometric\_Path\_Planner}(\mathcal{X} \subset X, \state_s, \state_f)$
\Statex $\tau \gets \texttt{HDDP}(\xi,X,U,F)$
\Statex $\tau^* \gets \texttt{CHDDP}(\tau,X, \mathcal{X}_{free},U,F)$
\Ensure $\tau^{*}$
\end{algorithmic}
\end{algorithm}

While DDP without clutter or hybrid dynamics is often capable of finding an optimal solution even from a bad initial trajectory, this is typically not true in the case of a hybrid system in clutter. It often gets stuck in a local minima that does not reach the goal, because  it fails to get away from an obstacle or to find the right sequence of modes. In fact, it is \textit{critical} that we start with a nominal trajectory that has minimal collision while accomplishing the task. 

Starting from a collision-free path\footnote{We refer to path as a sequence of states with no associated action. This path may not be kinodynamically feasible.} found by a path-planning algorithm such as visibility graph or RRT (mentioned subsequently), we first use HDDP to solve for a feasible state-action-mode sequence that closely tracks the reference path $\hat{\xi}  = \{\state_0, \state_1, \cdots \state_T\}$ such that every $\state_t$ along the path is at least $\epsilon$ away from any obstacle.  We do not impose collision avoidance constraints here, but as the reference path is at least $\epsilon$ away from any obstacle, the resulting state-action trajectory is likely to be collision free. 
\begin{equation*}
\begin{array}{rll}
\mathop {\min }\limits_{\control_0, \prob_0, \state_1, \control_1, \prob_1 \cdots,\state_T,\prob_T} & & \sum\limits_t {\phi({\state_t} - \hat \state_t) + |{\control_t}|_2^2} \\
\text{subject to} & & {\state_{t + 1}} = \hat f(\state_t,\control_t,\prob_t), \\
& & {\control}_t \in [\underline{\action_t},  \overline{\action_t}], \ \ \forall \ t.
\end{array}
\end{equation*}

Here $\phi(\cdot)$ denotes the soft-abs function (pseudo-Huber loss). We use the soft-abs function on the state deviation because its gradient will not diminish outside of a given range, so the optimization can continue to minimize the error until it is within the given range. We use a quadratic cost on the magnitude of the continuous action, since we prefer smaller $u_t$, but it does not have to be zero.

We use some path-planning algorithm to get an initial collision-free path. For problems that can be projected to two dimensions, we use a visibility graph planning method~\citep{lozano1979algorithm} to get multiple feasible paths
in different homotopy classes~\citep{bhattacharya2010search}. This heuristic provides efficient and good quality feasible paths
covering different subspaces of the search space, thereby also increasing the probability of a successful solution by CHDDP. We show in Figure \ref{fig:w5_hom} how this initialization method leads to multiple good quality CHDDP solutions. If CHDDP fails to generate a feasible solution with any geometrically feasible paths obtained from the homotopy stage, or if the path planning must happen in higher dimensional spaces, we use the probabilistically complete sampling-based RRT (Rapidly-exploring Random Trees) algorithm~\citep{lavalle2001randomized}.

\begin{figure}[th]
    \centering
    \includegraphics[width=\columnwidth]{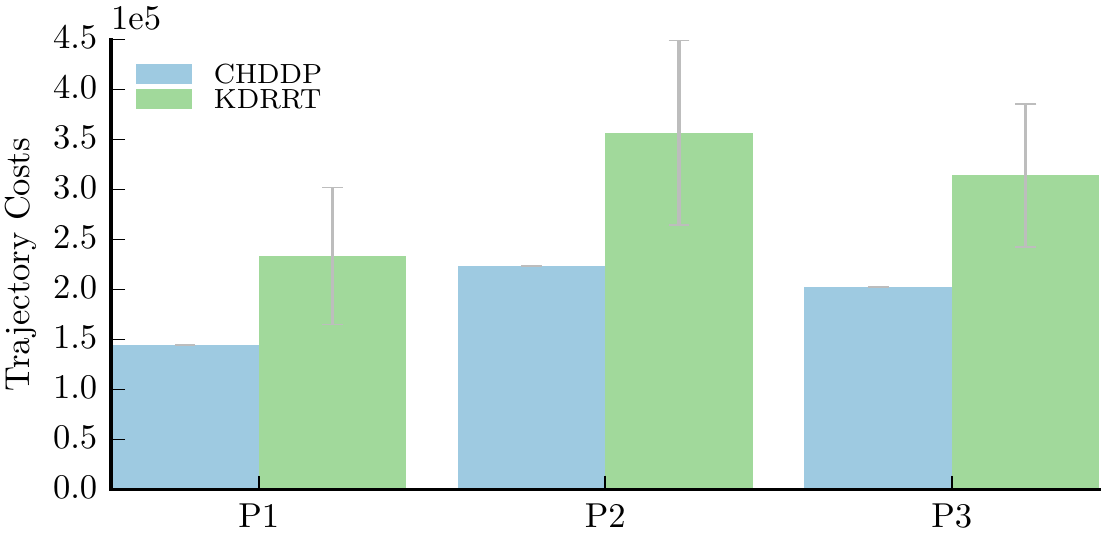}
    \caption{We compare the cost (according to our objective function) of the trajectories returned by CHDDP
    to those returned by KDRRT, for the three box pushing problems visualized in Figure~\ref{fig:pushing_hddp_rrt}.
    We average over 50 different runs of KDRRT for each problem (the mean and standard error bars are shown here).
    For the initialization of CHDDP, we choose the geometrically shortest homotopy path
    obtained from the visibility graph heuristic. This makes it deterministic, so we only run it once to compute the cost for each problem.
    }
    \label{fig:ddp_vs_rrt_costs}
\end{figure}

\subsection 
{Optimizing with a fixed mode sequence} 
After getting a feasible state-action trajectory, we use it to initialize the second optimization, where we optimize the original cost function with collision-free constraints:

\begin{align*}
    \min_{\control_0, \state_1, \cdots,\state_T}  &  &{l_T}(\state_T) &+ \sum_{t = 0}^{T - 1} {{l_{{a_t}}}({\state_t},{\control_t})} \\
 \text{subject to} &  &\state_{t+1} &= \hat f(\state_t, \control_t, \prob_t),\\
 & & & \state_t \in \mathcal{X}_{free}, \\
 & &  &{\control}_t \in [\underline{\action_t},  \overline{\action_t}], \, \, \forall \ t,
\end{align*}

where ${\prob_t}$ is obtained from Sub-problem I.
Our CHDDP algorithm modifies the HDDP method to handle the collision avoidance constraint explicitly. The first forward pass gives us a feasible trajectory (the initial trajectory). During the backward pass, we perform collision checking on the updated trajectory, and shrink the step size until the new trajectory is collision-free. Repeating this process, the forward passes will always produce feasible trajectories. As mentioned earlier, the collision checking is very efficient with IRIS regions as it effectively involves checking affine inequalities.

In this Sub-problem, we only update the trajectory in terms of $\{\state_t, \control_t\}$, while keeping the choice of mode $\prob_t$ unchanged. Otherwise, the trajectory would vary rapidly during the first few iterations and might converge to some different local minima, which could be really bad due to the existence of obstacles. Practically, we do this by using the cost coefficient $C_{ST}$ from the ending of Sub-problem I, which is already a large number.

We outline our overall algorithm for Hybrid DDP in Clutter (CHDDP) in Algorithm~\ref{alg:chddp}. After getting one or more geometrically feasible paths $\xi$, we run \texttt{HDDP} to get a feasible $\tau$ and then optimize this with \texttt{CHDDP}. 

\section{Experiments}
\label{sec:results}

\begin{figure*}[th]
    \begin{subfigure}[b]{0.33\textwidth}
    \centering
    \includegraphics[trim={2.5cm 1.2cm 2.5cm 0.8cm},clip,width=\textwidth]{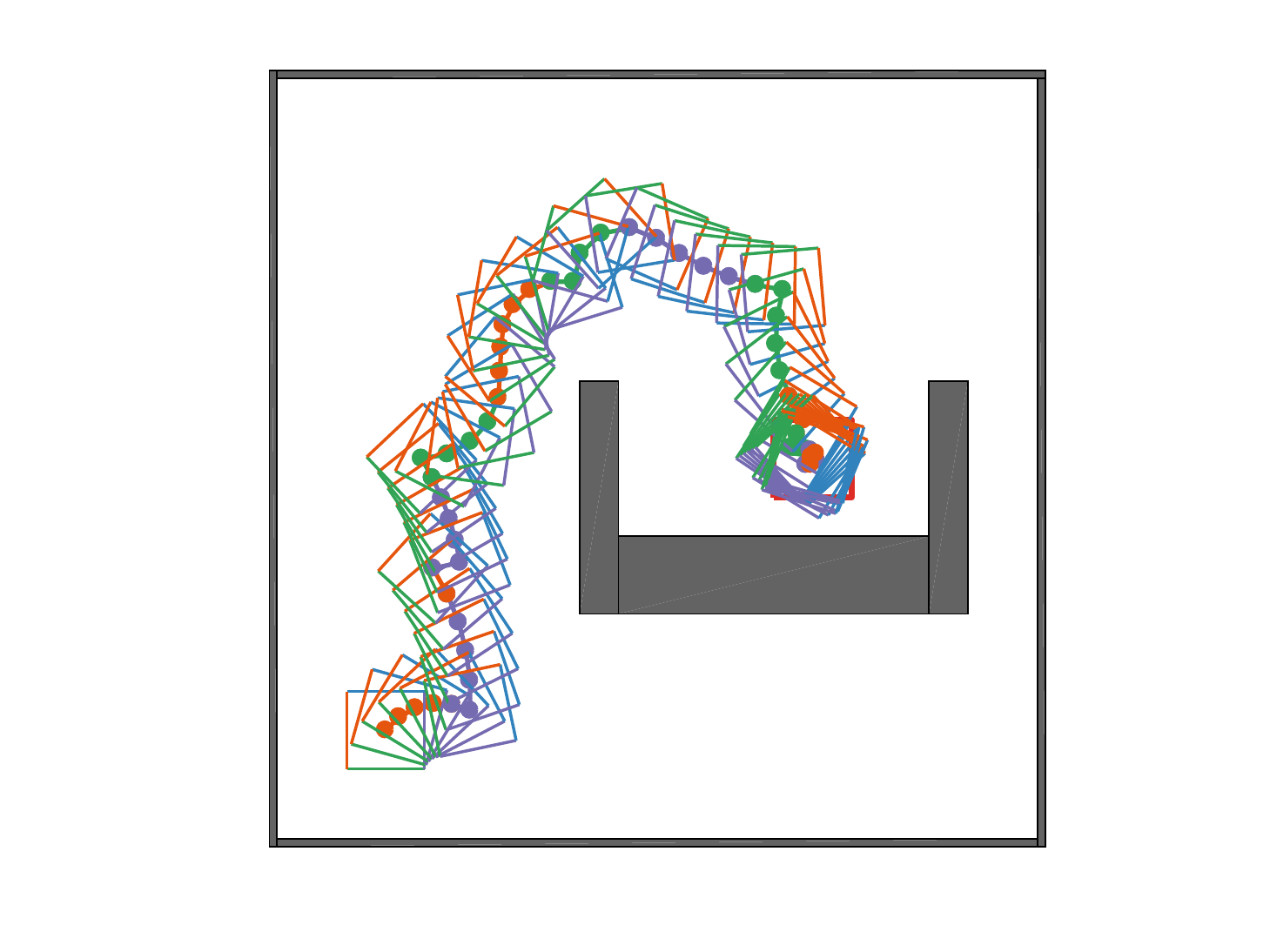}
    \caption{KDRRT}
    \label{fig:w1_rrt}
    \end{subfigure}
    \begin{subfigure}[b]{0.33\textwidth}
    \centering
    \includegraphics[trim={2.5cm 1.2cm 2.5cm 0.8cm},clip,width=\textwidth]{fig/pusher/w2_RRT.pdf}
    \caption{KDRRT}
    \label{fig:w2_RRT}
    \end{subfigure}
    \begin{subfigure}[b]{0.33\textwidth}
    \centering
    \includegraphics[trim={2.5cm 1.2cm 2.5cm 0.8cm},clip,width=\textwidth]{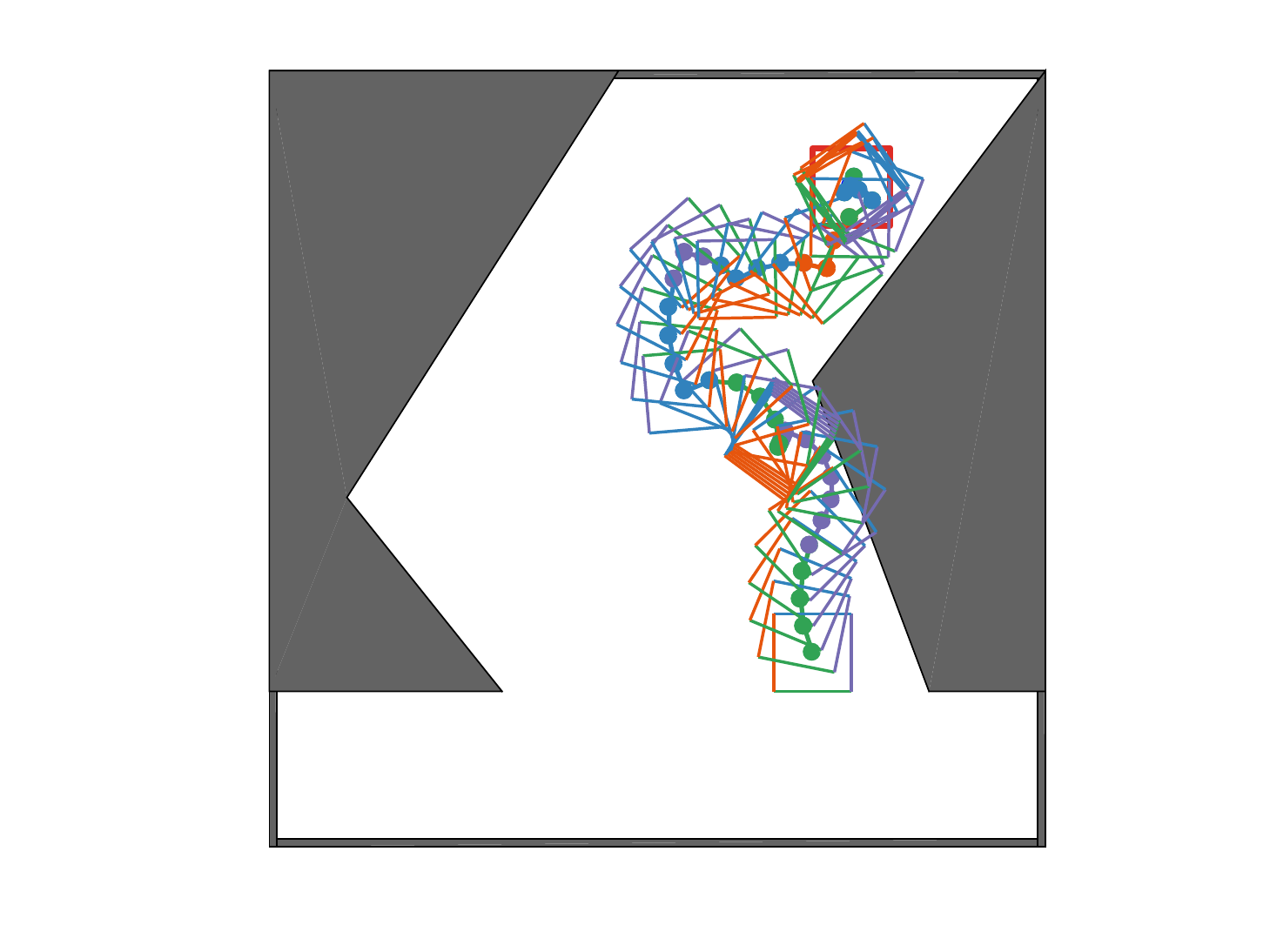}
    \caption{KDRRT}
    \label{fig:w3_rrt}
    \end{subfigure}
    \begin{subfigure}[b]{0.33\textwidth}
    \centering
    \includegraphics[trim={2.5cm 1.2cm 2.5cm 0.8cm},clip,width=\textwidth]{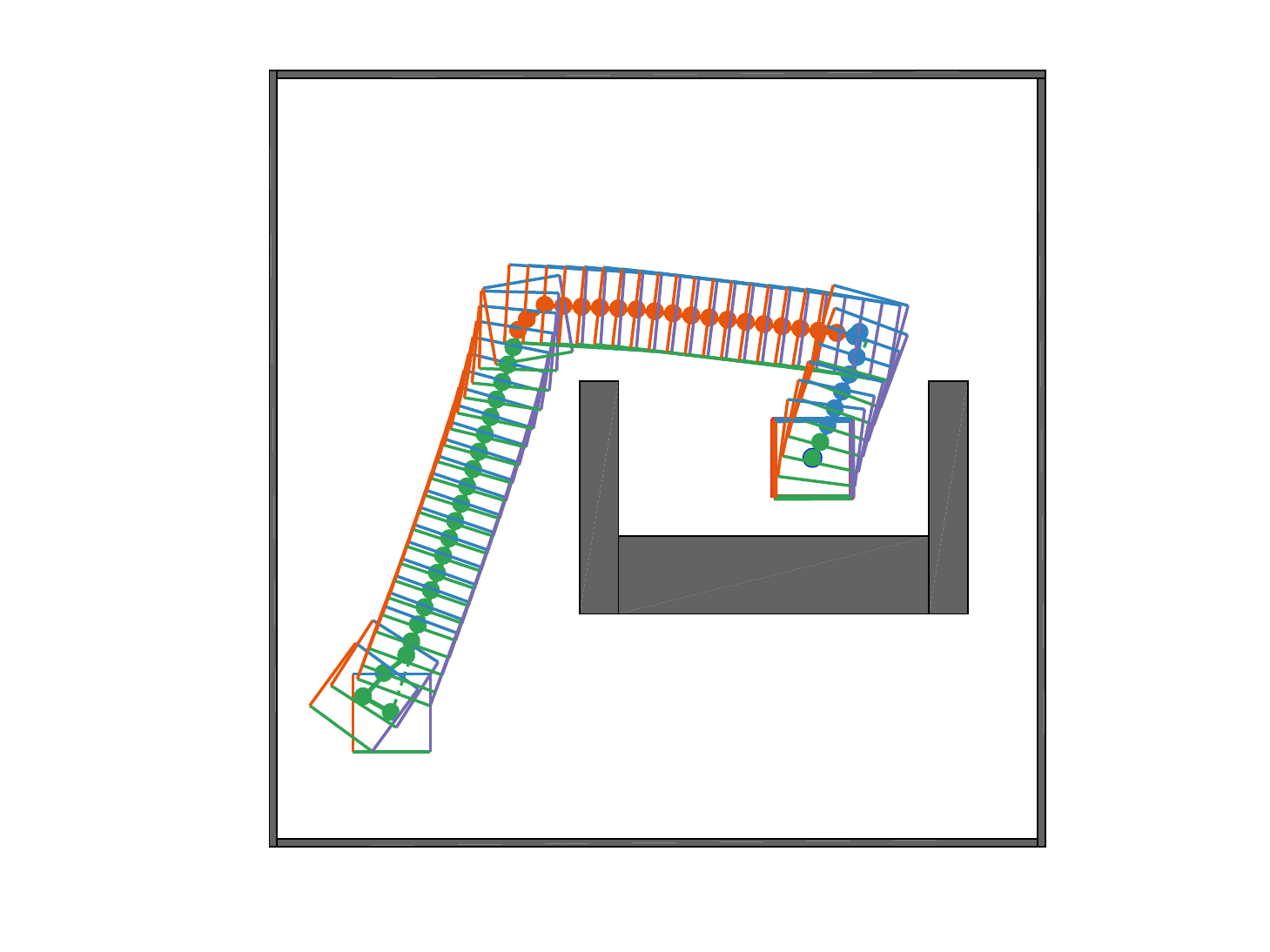}
    \caption{CHDDP}
    \label{fig:w1_hddp}
    \end{subfigure}
    \begin{subfigure}[b]{0.33\textwidth}
    \centering
    \includegraphics[trim={2.5cm 1.2cm 2.5cm 0.8cm},clip,width=\textwidth]{fig/pusher/w2_DDP.pdf}
    \caption{CHDDP}
    \label{fig:w2_HDDP}
    \end{subfigure}
    \begin{subfigure}[b]{0.33\textwidth}
    \centering
    \includegraphics[trim={2.5cm 1.2cm 2.5cm 0.8cm},clip,width=\textwidth]{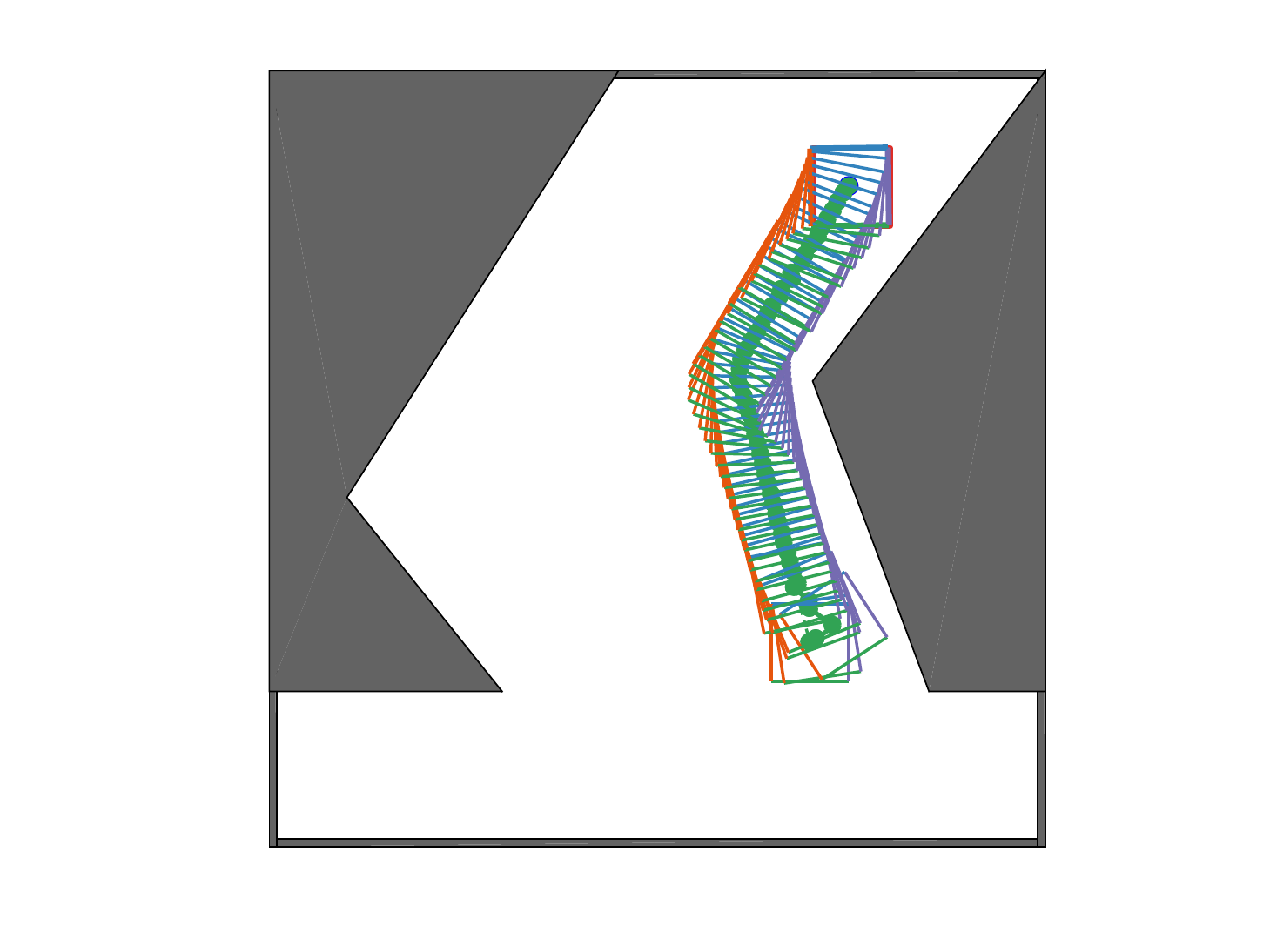}
    \caption{CHDDP}
    \label{fig:w3_hddp}
    \end{subfigure}
    \caption{
        CHDDP (lower row), initialized with the shortest homotopy path obtained from the visibility graph, 
        has significantly smoother trajectories than Kinodynamic RRT (upper row) for a range of box-pushing problems, both in terms of the quality and the number of mode switches. These three problems correspond in order to the ones in Figure~\ref{fig:ddp_vs_rrt_costs}.
    }
    \label{fig:pushing_hddp_rrt}
\end{figure*}

We demonstrate the effectiveness of our approach with the problem setting of pushing an unknown box. We use a similar setup to that of previous work our approach is based on~\citep{pajarinen2017hybrid} - please refer to them for details of the dynamics, control modes and the objective function. Here we only state the differences between our model and theirs, for the sake of brevity. The \emph{state-space} of our system is a subset of the 2D Special Euclidean Group (\emph{SE(2)}), and each state is represented as $\mathbf{x} = \left(x,y,\theta\right)$ where the co-ordinates refer to the centre of the box, and $\theta$ is the rotation of the box about the centre. We use a motion cone of $[-70\degree, 70\degree]$. Also, rather than having a continuous variation of the point being pushed, which would actually cause a tangent friction force, we push a single point on the edge at a time. 

The box pushing example is the only one that we run tests on, however it is equivalent to the popular Dubin's car model~\citep{dubins1957curves} with gears, another widely applicable hybrid control system.  Our experiments are on different sets of obstacles and start and goal poses for the box.

We have already shown in Figure~\ref{fig:w5_hom} how multiple geometrically feasible paths in various
homotopy classes can be used to initialize CHDDP for good quality locally optimal solutions.
For several additional problems, we compare our CHDDP against a baseline trajectory obtained from kinodynamic RRT with random sampling of modes and actions. We show quantitative results in Figure~\ref{fig:ddp_vs_rrt_costs}, where we compare the
costs (as per our objective function) of the trajectories returned by KDRRT and CHDDP. We average over 50 runs of KDRRT for each problem, while we deterministically initialize CHDDP with the geometrically shortest homotopy path, if more
than one exists for the problem. We also visualize some solution trajectories in Figure~\ref{fig:pushing_hddp_rrt}.
For all of the examples, the solutions produced by CHDDP are of significantly higher quality than those from KDRRT. Since CHDDP is the first attempt to have a locally optimal approach for this specific problem domain, we do not really have any other baseline method to directly compare our performance against. 

\begin{figure*}
    \centering
    \begin{subfigure}[b]{0.245\textwidth}
        \centering
        \includegraphics[trim={2.5cm 1.2cm 2.5cm 0.8cm},clip,width=\textwidth]{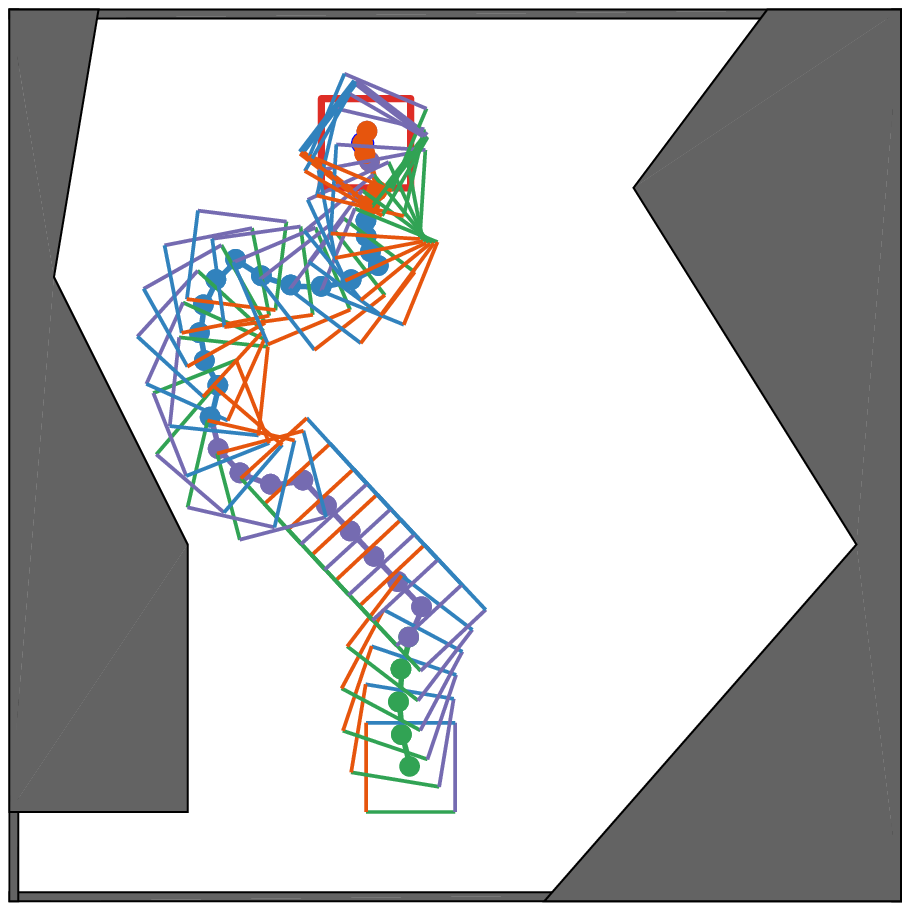}
        \caption{KDRRT only}
        \label{fig:p1_onlyrrt}
    \end{subfigure}
    \begin{subfigure}[b]{0.245\textwidth}
        \centering
        \includegraphics[trim={2.5cm 1.2cm 2.5cm 0.8cm},clip,width=\textwidth]{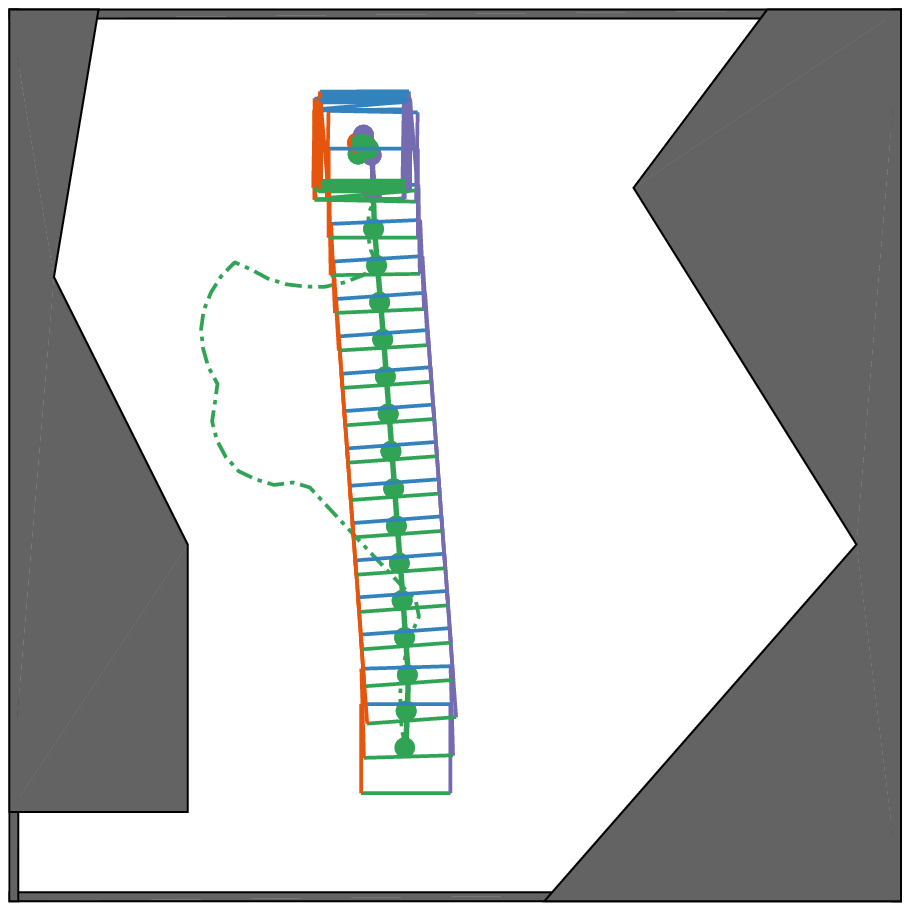}
        \caption{KDRRT + CHDDP}
        \label{fig:p1_rrt_chddp}
    \end{subfigure}
    \begin{subfigure}[b]{0.245\textwidth}
        \centering
        \includegraphics[trim={2.5cm 1.2cm 2.5cm 0.8cm},clip,width=\textwidth]{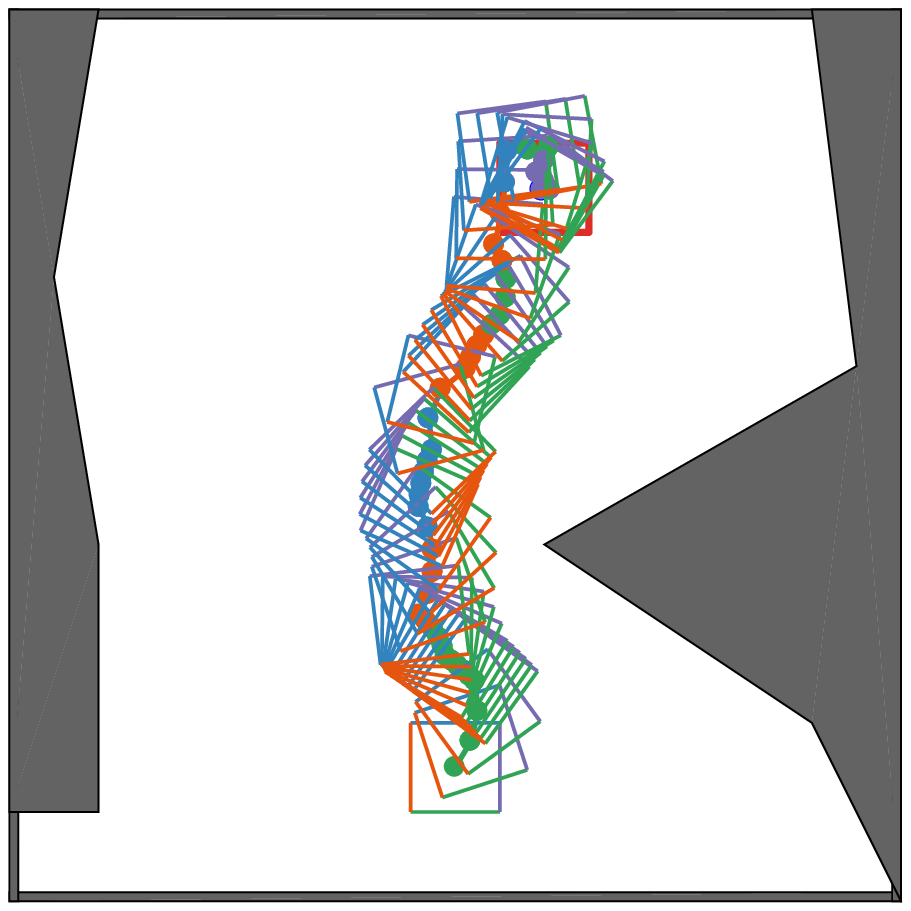}
        \caption{KDRRT only}
        \label{fig:p2_onlyrrt}
    \end{subfigure}
    \begin{subfigure}[b]{0.245\textwidth}
        \centering
        \includegraphics[trim={2.5cm 1.2cm 2.5cm 0.8cm},clip,width=\textwidth]{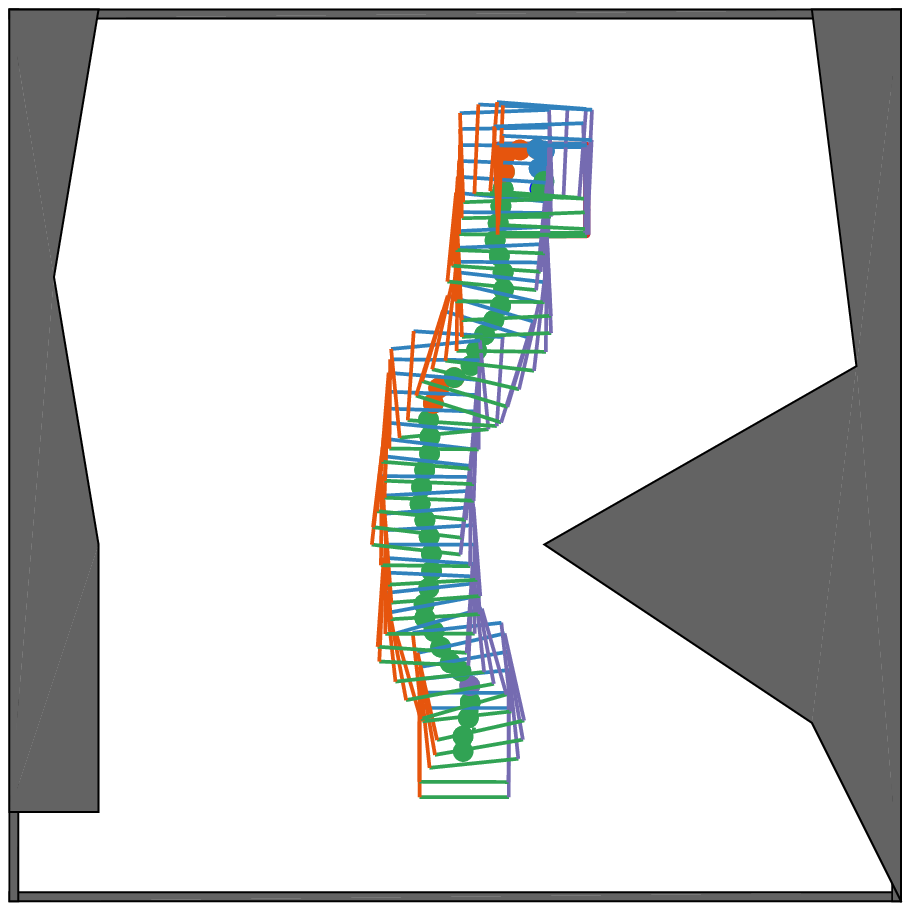}
        \caption{KDRRT + CHDDP}
        \label{fig:p2_rrt_chddp}
    \end{subfigure}
    \caption{Our CHDDP algorithm also works well when initialized with a KDRRT trajectory, which can be arbitrarily jagged. As shown in (\subref{fig:p1_rrt_chddp}) and (\subref{fig:p2_rrt_chddp}), the trajectories are much smoother than the
    corresponding initial KDRRT trajectories in (\subref{fig:p1_onlyrrt}) and (\subref{fig:p2_onlyrrt}).
    We do require that the initial trajectory maintain sufficient clearance from the obstacles, which we ensure
    via C-space obstacle inflation.
    }
    \label{fig:ddp_plus_kdrrt}
\end{figure*}

As we have mentioned before, the core of our approach is independent of the method used to generate
an initial solution. The visibility graph method for generating homotopy paths is a means of obtaining initial
feasible solutions that are short, while having good clearance (due to obstacle inflation).
However, should it not be applicable for the specific problem, 
CHDDP is also able to derive good quality trajectories and locally optimal
control policies when initialized by the Kinodynamic RRT. We show in Figure~\ref{fig:ddp_plus_kdrrt}
how even quite jagged and poor quality KDRRT trajectories are improved considerably
by CHDDP.

\section{Conclusion}
\label{sec:conclusion}

We proposed CHDDP, a method of efficiently obtaining locally optimal solutions for a hybrid control problem in a cluttered environment, which is a computationally hard problem. We do this by dividing the overall problem into feasible sub-problems. We approximate the feasible state space with a union of convex collision-free regions, which also makes collision checking much more efficient. For generating initial solutions for the hybrid DDP method, we generate high quality geometrically feasible paths in multiple homotopy classes. We solve the difficult hybrid DDP problem in two phases, each of which is more tractable than the joint problem. This allows us to extend the capability of recently proposed work to solve for hybrid control trajectories in the presence of obstacles. 

In our current approach, the hybrid DDP is used to generate the nominal state-action-mode trajectory without considering collision constraints. Therefore the feedback control gain does not take the collisions into account. This is a problem that could be addressed in potential future work.

\footnotesize{
\bibliographystyle{abbrv}
\bibliography{references}
}

\end{document}